\documentclass[10pt]{article} %
\usepackage[preprint]{tmlr}

\usepackage{amsmath,amsfonts,bm}

\def\eqref#1{equation~\ref{#1}}

\def\1{\bm{1}}

\DeclareMathAlphabet{\mathsfit}{\encodingdefault}{\sfdefault}{m}{sl}
\SetMathAlphabet{\mathsfit}{bold}{\encodingdefault}{\sfdefault}{bx}{n}

\newcommand{\simulator}{\mathcal{S}}
\newcommand{\ztrainset}{\mathcal{Z}_\text{train}}
\newcommand{\ztestset}{\mathcal{Z}_\text{test}}

\newcommand{\ztest}{z_\text{test}}
\newcommand{\rpastset}{\mathcal{R}_\text{past}}
\newcommand{\rfutureset}{\mathcal{R}_\text{future}}
\newcommand{\hatloss}{\hat{L}}
\newcommand{\infscore}{\mathcal{I}(z, z_\text{test})}

\newcommand{\obsteps}{\mathcal{T}}
\newcommand{\simflinear}{\textsc{Simfluence-Linear}}
\newcommand{\simfadd}{\textsc{Simfluence-Additive}}
\newcommand{\simfmult}{\textsc{Simfluence-Multiplicative}}
\newcommand{\tracin}{\textsc{TracIn}}
\newcommand{\tracincp}{\textsc{TracIn-CP}}
\newcommand{\tracinideal}{\textsc{TracIn-Ideal}}

\usepackage{hyperref}
\usepackage{url}

\title{Simfluence: Modeling the Influence of Individual \\ Training Examples by Simulating Training Runs}

\author{
\begin{center}

Kelvin Guu$^{*,1}$ \quad Albert Webson$^{*,\diamond,2}$ \quad Ellie Pavlick$^{1,2}$ \quad Lucas Dixon$^1$
\AND
Ian Tenney$^1$ \quad Tolga Bolukbasi\thanks{Lead contributors. Please see Contributions section for details. $^\diamond$ Work done during an internship at Google Research.}$\enskip^{,1}$
\AND
\small{\texttt{\{kguu,epavlick,ldixon,iftenney,tolgab\}@google.com}}, $^1$Google Research
\\
\small{\texttt{\{albert\_webson,ellie\_pavlick\}@brown.edu}}, $^2$Brown University

\end{center}
}

\def\month{MM}  %
\def\year{YYYY} %
\def\openreview{\url{https://openreview.net/forum?id=XXXX}} %

\usepackage{booktabs}
\usepackage{multirow}
\usepackage{caption}
\usepackage{subcaption}
\usepackage{graphicx}
\graphicspath{{./raster/}}

\begin{document}

\maketitle

\begin{abstract}

Training data attribution (TDA) methods offer to trace a model's prediction on any given example back to specific influential training examples. Existing approaches do so by assigning a scalar \emph{influence score} to each training example, under a simplifying assumption that influence is additive, whereby the total influence of a training set is the sum of its parts. But in reality, we observe that training examples interact in highly non-additive ways due to factors such as inter-example redundancy, training order, and curriculum learning effects.

To study such interactions, we propose \emph{Simfluence}, a new paradigm for TDA where the goal is not to produce a single influence score per example, but instead a \emph{training run simulator}: the user asks, \emph{``If my model had trained on example $z_1$, then $z_2$, ..., then $z_n$, how would it behave on $\ztest$?''}; the simulator should then output a \emph{simulated training run}, which is a time series predicting the loss on $\ztest$ at every step of the simulated run. This enables users to answer counterfactual questions about what their model would have learned under different training curricula, and to directly see where in training that learning would occur.

Under the Simfluence paradigm, we present a simulator (\simflinear{}) that captures important non-additive interactions using a Markov process. It is often able to predict the spiky trajectory of individual example losses with surprising fidelity, while matching the interpretability of prior TDA work and running in milliseconds. Furthermore, we show that existing TDA methods such as \tracin{} and influence functions can be viewed as special cases of \simflinear{}. This enables us to directly compare methods in terms of their simulation accuracy, subsuming several prior TDA approaches to evaluation. In experiments on large language model (LLM) fine-tuning, we show that our method predicts loss trajectories with much higher accuracy than existing TDA methods (doubling Spearman's correlation and reducing mean-squared error by 75\%) across several tasks, models, and training methods.

\end{abstract}

\begin{figure}[t!]
     \centering
     \begin{subfigure}[b]{0.325\textwidth}
        \centering
        \includegraphics[width=\textwidth]{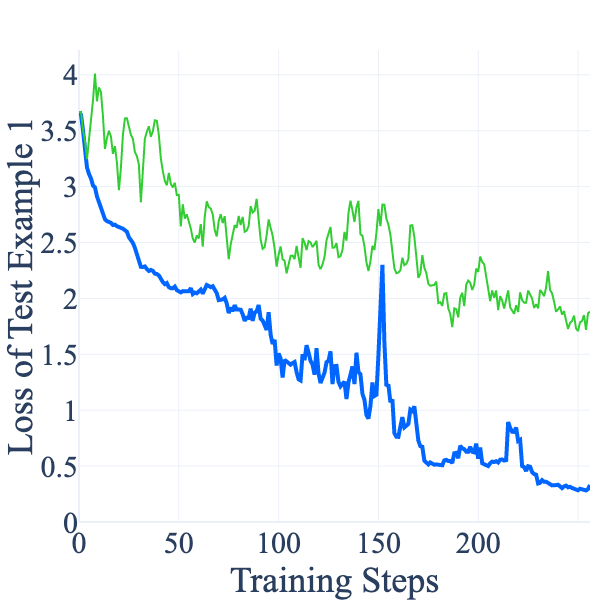}
     \end{subfigure}
     \hfill
     \begin{subfigure}[b]{0.325\textwidth}
        \centering
        \includegraphics[width=\textwidth]{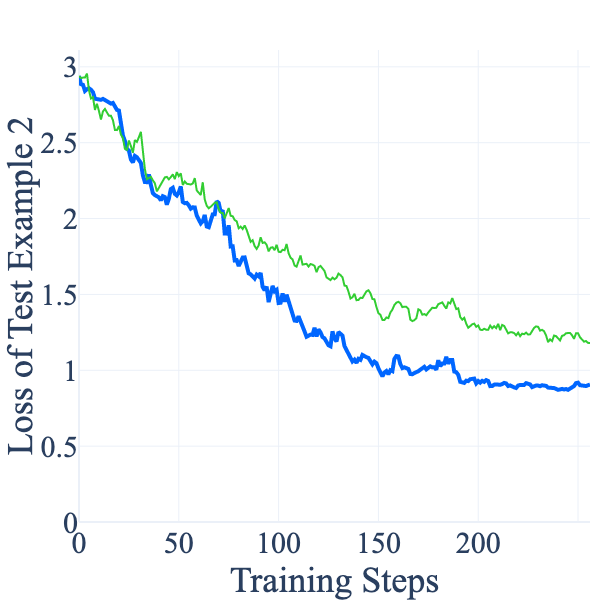}
     \end{subfigure}
    \hfill 
    \begin{subfigure}[b]{0.325\textwidth}
        \centering
        \includegraphics[width=\textwidth]{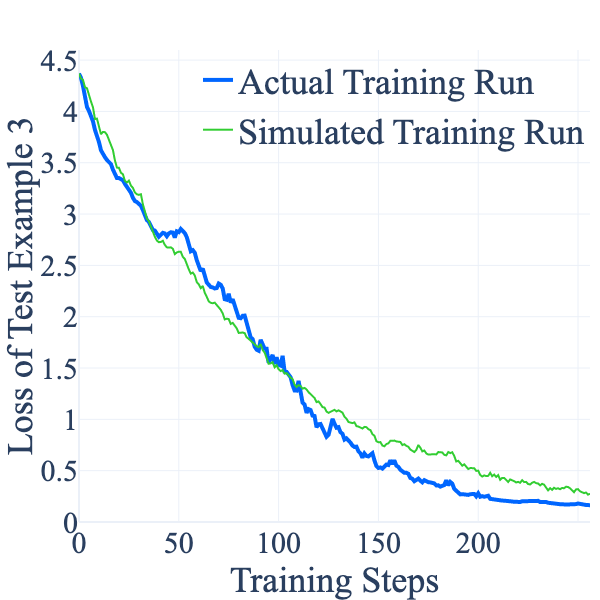}
     \end{subfigure}
     \caption{
     Training data attribution (TDA) methods seek to understand the effect of individual training examples.
     Simfluence is a new paradigm for TDA, where the goal is to develop \emph{training run simulators} that can accurately predict how any given sequence of training examples would affect the model's loss on any particular test example. Here, we plot the loss of three different test examples over the course of a training run. We compare the true observed loss trajectories (blue) with our simulator's predicted trajectories (green).
     Surprisingly, many of the ups and downs in the true loss trajectories are not ``random'' but can be anticipated by our simulator, showing the extent to which our simulator understands the effect of each training example.}
     \label{fig:motivation}
\end{figure}

\section{Introduction}\label{sec:intro}

Important advances in machine learning are often made possible by better or more data \citep{halevy2009unreasonable, deng2009imagenet, kaplan2020scaling}. But which training examples are actually responsible for specific model successes or failures? Training data attribution (TDA) methods seek to answer this question.

Many existing TDA methods share the following formulation: given any test example of interest, $\ztest$, they aim to identify which training examples caused the biggest change in the model's loss on $\ztest$. If a training example $z$ helped reduce the loss on $\ztest$, it is called a \emph{proponent}, while increasing the loss makes it an \emph{opponent}. The amount of change in loss on $\ztest$ is called the \emph{influence} of $z$ on $\ztest$, which we denote $\infscore$. This description encompasses some of the most well-known methods for TDA, such as influence functions \citep{Koh,hampel1974influence,cook1980characterizations} and \tracin{} \citep{Pruthi}. Both aforementioned methods make a simplifying assumption that influence is \emph{additive}: the combined influence of several examples should equal the sum of their individual influence scores. Influence functions make this assumption within a local neighborhood of the model's parameters, while \tracin{} assumes this when integrating over training steps. But this common assumption does not adequately capture important non-additive aspects of real training, which we discuss next.

\paragraph{Non-additive effects.} Many examples in a training set may provide similar or redundant information. This is generally hard to express when using a single influence score per training example. For example, influence functions \citep{Koh} evenly ``split the credit'' among all redundant examples, meaning that 100 examples teaching a very common but essential piece of information could all rank lower than one example teaching something less important but unique. Likewise, \tracinideal{} \citep{Pruthi} tends to handle the issue by assigning credit to whichever example arbitrarily appeared earlier in training --- hiding the otherwise equivalent value of later examples. These limitations were noted by \cite{sogaard2021revisiting}, which found that existing TDA methods tend to over- or underestimate influence in such situations, as well as by \cite{koh2019accuracy} who found that influence systematically underestimated the effect of group interventions. Redundancy between two examples ($z_1$ and $z_2$) is a special case of \emph{submodular interaction} \citep{fujishige2005submodular}: informally speaking, $\text{information}(z_1 \cup z_2) < \text{information}(z_1) + \text{information}(z_2)$. Conversely, if something can only be learned through the combination of multiple examples (e.g., as observed in curriculum learning \citep{bengio2009curriculum}), then we have a \emph{supermodular interaction}, which also is not modeled by existing TDA methods.

\paragraph{Influence as counterfactual simulation.} To account for these important phenomena, we propose \emph{Simfluence}, a new paradigm for TDA where the output is a \emph{training run simulator}, rather than a single score per example. In this setup, the user is interested in the model's performance on some example $\ztest$, and they would like to simulate various counterfactual training scenarios, such as: ``What if I had removed \emph{<X>} group of examples from my training data?'' \citep{koh2019accuracy}, or ``What if I had trained on \emph{<Y>} first, then \emph{<X>}?'' \citep{sogaard2021revisiting}, or ``What if I duplicated \emph{<Z>} ten times?'' \citep{han2022orca}.

To explore these questions, the user first poses a training curriculum (which examples are seen, and in what order). Then, the simulator generates a time series that predicts what the loss on $\ztest$ would be after each step of the training run (a \emph{loss trajectory}). Such a loss trajectory can directly show the user which training steps are helping or hurting, and what the final loss will be. It can also change depending on the order and combination of training examples, thereby modeling non-additive effects. Finally, the accuracy of this simulated trajectory can be directly validated by comparing it to the true loss trajectory during an actual training run using the same curriculum; for instance, Figure \ref{fig:motivation} shows several examples of how well our simulator can predict true trajectories.

Inspired by the framing of simulation, we propose a new simulator, \simflinear{}, that models non-additive effects caused by training order and redundancy using a Markov process, while preserving the interpretability of prior TDA methods. Furthermore, we show that both \tracin{} and influence functions can be reproduced as special cases of this simulator. By framing both prior and proposed methods as simulators, we are now able to directly compare them, evaluating each in terms of their simulation accuracy against real loss trajectories. We find that \simflinear{} significantly outperforms existing methods (doubling Spearman's correlation and reducing mean squared error by 75\%) across several tasks and models in the setting of large language model (LLM) fine-tuning (both standard full-model tuning and parameter-efficient tuning).

\section{Task}\label{sec:task}

Here, we formally define Simfluence, the task of training run simulation. Let $z$ denote a single example. Let $\ztrainset$ and $\ztestset$ denote the set of all training examples and test examples\footnote{Or validation examples, depending on the use case.}, respectively. We use $i=1,\dots,n$ to index over training examples. We use $c = (c_1, c_2, \dots, c_t, \dots, c_T)$ to specify a training curriculum: for each training step $t=1,\dots,T$, let $c_t \subset \{1,\dots,n\}$ be a set of integers specifying the batch of training examples consumed at step $t$. Then, let $L_t(z)$ denote the loss on example $z$ after taking training step $t$.

A training run simulator, $\simulator$, should be able to simulate the loss of any given test example, $\ztest$, over the course of training on any arbitrary curriculum. Formally, it takes two inputs: 1) the curriculum, $c$, and 2) the initial loss before training begins, $L_0(\ztest)$. It then outputs a predicted loss, $\hat{L}_t(\ztest)$, for every time step (the loss trajectory):
\begin{equation*}
\simulator(c, L_0(\ztest)) \mapsto \hat{L}_1(\ztest), \hat{L}_2(\ztest) \dots, \hat{L}_T(\ztest)
\end{equation*}
We are mainly interested in how this loss trajectory would change as a function of the curriculum, $c$.

Finally, we note that there is generally a trade-off between simulation accuracy and simulation speed. At one extreme, we could always obtain a perfect simulation of training by just \emph{actually running training}, but this is generally too slow for quickly probing and exploring the effect of different training examples and curricula. A useful simulator should be orders of magnitude faster to run; ideally taking seconds instead of hours or days. Whenever possible, we also prefer simulators with interpretable structure, provided this does not compromise simulation accuracy.

\subsection{Training a simulator}

In this work, we consider simulators that are themselves also \emph{learned models}. We will learn our simulator from previously conducted training runs. For each previous run, we record what curriculum was used ($c$, the input to our simulator), and the true observed loss trajectory of a given example $\ztest$ (which we denote $L_{1:T}(\ztest)$, which is the desired output of the simulator). Hence, each run provides an (input, output) pair for us to train a simulator via supervised learning. To formalize this, we will define a run as $r = (c, L_{1:T}(\ztest))$.

Once we have trained the simulator, we can then use it to predict loss trajectories for future planned runs, where we know what curriculum we will use, but have not observed the true loss trajectory. We will use $\rpastset$ to denote the set of previously observed runs, and $\rfutureset$ to denote future planned runs. For each run in $\rpastset$, we assume that we have recorded the loss for any test example of interest ($\ztest$) for some subset of time steps.

These requirements can be readily met in many practical scenarios. Model developers will often re-train their model multiple times before obtaining a version that meets their goals, and usually already track key metrics such as validation set losses over the course of each run. Hence, it often does not add much marginal cost to conduct a few extra runs for the purpose of learning a simulator.\footnote{Large language model pre-training is currently an exception to this, as even single training runs may require months.} Early work in TDA often assumed access to only one run \citep{Pruthi}, or even just the final checkpoint of one run \citep{Koh}, while more recent work has obtained better results by leveraging multiple runs \citep{sogaard2021revisiting}, which we also demonstrate in our experiments. 

\subsection{Evaluating a simulator}
\label{sec:eval-metrics}

To evaluate a simulator, $\simulator$, we can check how well its simulation of a future run, $r \in \rfutureset$, predicts what happens when we actually perform that training run, by comparing our simulator's predicted losses at each time step, $\hatloss_{1:T}(\ztest) = [\hatloss_1(\ztest),\dots,\hatloss_T(\ztest)]$, with the true observed losses, $L_{1:T}(\ztest) = [L_1(\ztest),\dots,L_T(\ztest)]$. We measure this using the mean squared error (MSE), which is averaged over test examples and training steps:
\begin{equation*}
\text{MSE}(\simulator) = \frac{1}{\ztestset} \frac{1}{T} \sum_{z \in \ztestset} \sum_{t=1}^T (L_t(z) - \hatloss_t(z))^2
\end{equation*}
Second, we wish to evaluate some simulators that do not provide a good absolute prediction of an example's loss, but may predict a good relative ordering of final losses among test examples at time $T$. For this, we use Spearman's correlation, which is only sensitive to the ranking among losses:
\begin{equation*}
\text{Spearman}_\text{final}(\simulator) = \text{Spearman}(\{(\hatloss_T(z), L_T(z))\text{ for }z \in \ztestset\})
\end{equation*}

By casting TDA as a simulation task, we now have an unambiguous goal and ground-truth: all simulators should strive to match real observed training runs. In Section~\ref{sec:connections}, we show how existing TDA methods can also be formulated as simulators, enabling us to directly compare our approach and prior work all by the same yardstick.

In contrast, prior TDA research has not always agreed on a shared ground-truth or evaluation setup. For example, \cite{Koh} used leave-one-out retraining (LOO) as their ground-truth. In this setup, a practitioner removes exactly one training example $z$ from the training set and then re-trains their model. They then evaluate whether the resulting change in loss on a test example $\ztest$ correlates well with the predicted influence score, $\infscore$. This can be considered a special case of our simulation task, where $\rpastset$ and $\rfutureset$ differ by only one example. Recent work by \cite{sogaard2021revisiting} has argued that LOO retraining tends to result in small or negligible changes in loss (due to redundancy among training examples), and hence argue that it is not the most important quantity that all TDA methods should aspire to predict.

\section{Approach}\label{sec:approach}

\subsection{Simulator}
We now present a simple training run simulator which models the loss trajectory of a single test example, $z$, as a \emph{linear Markov process}. Given the loss at time $t-1$, we model the loss at time $t$ to be:
\begin{equation} \label{eq:model}
L_{t}(z) = {\alpha(c_t)} L_{t-1}(z) + {\beta(c_t)}
\end{equation}

where multiplicative factor $\alpha(c_t)$ and additive factor $\beta(c_t)$ are learned functions of $c_t$ (the batch of training examples consumed at step $t$). We call this model \emph{\simflinear{}}.

This model expresses our simplifying assumption that the change in loss at step $t$ is a linear function of the training examples consumed at that step. For example, if $\alpha(c_t) = 0.5$ and $\beta(c_t) = -2.1$, this would mean that the training examples at step $t$ cause a 50\% relative reduction in the loss, followed by a 2.1 absolute reduction.

Performing simulation with this model is straightforward. We start with the known initial loss, $L_0(z)$. We then recursively apply Equation~\ref{eq:model} to predict subsequent time steps: from $L_0(z)$ we predict $L_1(z)$, which is then used to predict $L_2(z)$, and so on.

Next, we elaborate on the learned functions $\alpha(c_t)$ and $\beta(c_t)$. Recall that $c_t \subset \{1, \dots, n\}$ is a set of integers indicating the batch of training examples consumed at step $t$, and let the learnable parameters of \simflinear{} be $\Omega = (A, B)$, where $A \in \mathbb{R}^{n}$ and $B \in \mathbb{R}^{n}$. Then, we define $\alpha(c_t)$ and $\beta(c_t)$ as:
\begin{equation*}
\alpha(c_t) = \sum_{i \in c_t} A_{i} \qquad\quad
\beta(c_t) = \sum_{i \in c_t} B_{i}
\end{equation*}
Under this model, $A_{i}$ represents how much training example $z_i$ multiplicatively reduces the loss on test example $z$ --- we call this \emph{multiplicative influence}. Likewise, $B_{i}$ reflects the \emph{additive influence} of $z_i$ on $z$.

Note that we have two learned parameters ($A_{i}$ and $B_{i}$) for each training example, meaning that the entire simulator has $2n$ parameters total. Also, note that this simulator only models the loss of a single test example, $z$. To model $m$ test examples, we would learn $m$ separate simulators, each with $2n$ parameters. In Section~\ref{sec:limitations}, we discuss future work that could be more parameter-efficient.

In our experiments, we wish to study the relative importance of multiplicative and additive influence. To do so, we introduce two ablations of \simflinear{}:
\begin{itemize}
    \item \simfadd{}: we only model additive influence, disabling multiplicative influence by setting $\alpha(c_t) = 1$ for all $c_t$.
    \item \simfmult{}: we only model multiplicative influence, disabling additive influence by setting $\beta(c_t) = 0$ for all $c_t$.
\end{itemize}

In Section~\ref{sec:connections}, we will show that prior TDA methods can be viewed as special cases of \simfadd{}: they model additive influence, but not multiplicative influence.

The multiplicative factor $\alpha(c_t)$ in \simflinear{} is important because it enables us to model redundancy between training examples, and the effect of training order. For example, consider two training examples, $z_{1}$ and $z_{2}$, which both have the same multiplicative influence: $A_{1} = A_{2} = 0.5$. For simplicity, let us also assume they both have 0 additive influence, $B_{1} = B_{2} = 0$. Under this model, both examples have the same effect: each time we encounter them, they cut the loss in half. Now, suppose the initial loss is $L_0(\ztest) = 100$. If we take a training step on $z_1$ first, then our simulator predicts $z_1$ will reduce the loss from 100 to 50. If we then take a training step on $z_2$, it will further reduce the loss from 50 to 25. Both examples reduce the loss by 50\%, but the second example causes less absolute loss reduction (25 rather than 50), because it came second. If the examples were reversed, the reverse would be true. Such a phenomenon cannot be modeled using additive influence, which always assumes the same effect regardless of order.

\subsection{Learning the simulator}\label{sec:learn_simulator}

Let $\obsteps$ denote the subset of training steps in $\rpastset$ where we recorded the loss of test example $z$ (both \emph{before} and \emph{after} the training step). To learn the parameters of the simulator, $\Omega = (A, B)$, we simply minimize the following L2-regularized regression objective:
\begin{align}\label{eq:objective}
\mathcal{O}(A, B) &= \sum_{t \in \obsteps} \left(L_t(z) - \hatloss_t(z)\right)^2 + \lambda (\|A\|^2 + \|B\|^2) \\
\hatloss_t(z) &= \alpha(c_t) L_{t-1}(z) + \beta(c_t)
\end{align}
where $\lambda$ is a hyperparameter controlling the amount of L2 regularization. In Appendix~\ref{sec:objective_soln}, we show that this reduces to a standard multivariate linear regression problem, and provide the closed form solution.

When the training batch size is 1, we can simplify further. Let $\obsteps_i$ denote the subset of training steps where training example $z_i$ was encountered. Then, we can reduce $\mathcal{O}(A, B)$ to only the terms involving $z_i$:
\begin{equation}\label{eq:bs1_objective}
\mathcal{O}(A_i, B_i) = \sum_{t \in \obsteps_i} (L_t(z) - A_{i} L_{t-1}(z) - B_{i})^2 + \lambda (A_i^2 + B_i^2)
\end{equation}
This objective is now a \emph{univariate} linear regression problem (again with a closed form solution), which only depends on data from the time steps where example $z_i$ was observed, and has no dependence on any other parameters $A_{j}$ or $B_{j}$ for $j \neq i$. This is convenient: for each example $z_i$ of interest, we can estimate $A_i$ and $B_i$ without modeling any other examples.

\subsection{Data requirements}\label{sec:data_requirements}

We now analyze how much data is needed to learn \simflinear{}. For the sake of building intuition, we start by deriving the bare minimum amount of data needed for \simflinear{}'s training objective to have a unique solution, for the simple case where the training batch size is 1 and L2-regularization is disabled ($\lambda = 0$). As noted in the previous section, we need to solve Equation~\ref{eq:bs1_objective} for each training example of interest. Since Equation~\ref{eq:bs1_objective} is a univariate linear regression problem with two parameters ($A_i$ and $B_i$), it is necessary and sufficient to observe two training steps on $z_i$ for Equation~\ref{eq:bs1_objective} to have a unique solution. If we apply this requirement to all training examples, then we need $\rpastset$ to include at least two training steps for every training example of interest. This requirement can be met by a single training run of two epochs, or two training runs of one epoch each. For the more general case of batch size > 1, a similar conclusion holds: under reasonable conditions, we need roughly $2n$ training steps to estimate $2n$ parameters --- see Appendix~\ref{sec:sample_efficiency} for details and important requirements. In some settings, it may be computationally undesirable to compute losses after every training step. If so, we could still meet the requirement by only computing losses every $H$ steps, while doing $H$ times more runs or epochs to meet the overall $2n$ data requirement.

Finally, note that $2n$ is the bare minimum needed for a unique solution. We present this result mainly to show that data requirements scale linearly with $n$. In our experiments, we find that $20n$ to $60n$ training steps are more than sufficient. Note that we can always obtain more data for Simfluence by performing more training runs --- hence, the real limiting factor is the computational cost of training runs. Prior TDA methods were not formulated as learned models and therefore do not have a ``data requirement'' per se. However, we can compare their computational cost with \simflinear{} --- see Appendix~\ref{sec:compute} for details.

\subsection{Generality beyond gradient-based learning}

So far, we have presented \simflinear{} as a method to simulate gradient descent on a loss function. But it is worth noting that \simflinear{} contains no assumptions that are specific to gradient-based learning: unlike most other TDA methods, it does not require any access to model gradients or model parameters; it simply needs to know what examples are consumed at each step, and how losses change. Furthermore, instead of predicting test example losses, \simflinear{} could just as well be used to predict the trajectory of any arbitrary metric, such as the model's average accuracy across multiple examples, the L2-norm of a model's parameters, the gap between train and test loss, etc.

In the most generic terms, \simflinear{} is designed to simulate algorithms that incrementally consume examples, and to determine how those examples affect any given metric. This description encompasses numerous other algorithms of interest, such as in-context learning \citep{brown2020language} (where examples are consumed by the the forward pass of a large language model) or reinforcement learning (where the RL agent consumes ``episodes'', and the metric to track is expected reward). We hope to explore these applications in future work.

This framing also connects the Simfluence paradigm to research on credit assignment, such as Shapley values \citep{shapley1953value}. However, unlike the general credit assignment problem, we make the useful extra assumption that the metric of interest can be measured after each step of the process, rather than just the end.

\section{Connections to prior TDA methods}\label{sec:connections}

Here, we show how prior TDA methods, \tracin{} \citep{Pruthi}, and influence functions \citep{Koh}, can all be viewed as simulators under the Simfluence paradigm. In particular, we show that they are all special cases of \simfadd{} as they do not model multiplicative influence.

\subsection{\tracinideal{}}
\tracinideal{} was first introduced by \cite{Pruthi}. It assumes that we have performed a single training run, where each step consumes a single example (batch size 1). \tracinideal{} measures the net amount of loss reduction caused by a particular example $z_i$:
\begin{equation}\label{eq:tracinideal}
\mathcal{I}_\text{TracIn-Ideal}(z_i, z) = \sum_{t \in \obsteps_i} L_{t-1}(z) - L_{t}(z)
\end{equation}
where $\obsteps_i$ denotes the subset of time steps where example $z_i$ was encountered.

Here, we will show that \tracinideal{} \textbf{is equivalent to} \simfadd{}, a simple simulator that only models additive influence:
\begin{equation*}
L_{t}(z) = L_{t-1}(z) + \beta(c_t) \qquad \beta(c_t) = \sum_{i \in c_t} B_i
\end{equation*}

Let us consider the training objective for \simfadd{} when batch size equals 1 and L2 regularization is disabled ($\lambda = 0$). We start with Equation~\ref{eq:bs1_objective} and remove the multiplicative influence terms to get:
\begin{equation*}
\mathcal{O}(B_i) = \sum_{t \in \obsteps_i} (L_t(z) - L_{t-1}(z) - B_{i})^2
\end{equation*}
This objective has a simple closed-form solution --- $B_i$ should equal the negative average of all loss reductions where training example $z_i$ was consumed:
\begin{equation}\label{eq:simfadd_soln}
\hat{B}_{i} = -\frac{1}{\obsteps_i} \sum_{t \in \obsteps_i} L_{t-1}(z) - L_{t}(z) 
\end{equation}
We see that $\hat{B}_{i}$ is equivalent to the influence score defined by \tracinideal, up to a normalization term ($-1/\obsteps_i$), which is constant when all examples are encountered equally often. Hence, \tracinideal{} can be viewed as a simple simulator that only models the additive influence of $z_i$. 

Note that \tracinideal{} only uses data from a single training run. \cite{sogaard2021revisiting} proposed \emph{Expected \tracinideal{}}, where the \tracinideal{} score is averaged over multiple runs, rather than just one. This too is  encompassed by \simfadd{}, corresponding to the case where we have $\rpastset > 1$.

\subsection{\tracincp{}}\label{sec:tracin-cp}

 \tracincp{} (``CP'' stands for ``checkpoint'') was proposed by \citet{Pruthi} as a computationally cheaper approximation to \tracinideal. \tracincp{} makes two approximations. First, instead of summing loss reductions over all steps where training example $z_i$ is encountered ($\obsteps_i$), it only sums over steps where model checkpoints are saved, which we denote $\obsteps_\text{CP}$.

Second, it replaces the actual observed loss reduction at each step $t \in \obsteps_\text{CP}$ with the following approximation:
\begin{equation}\label{eq:tracincp_approx}
L_{t}(z) - L_{t+1}(z) \approx\eta_t \nabla_{\theta} L_{t}(z_i)^\top \nabla_{\theta} L_{t}(z)
\end{equation}
where $\eta_t$ is the learning rate at step $t$ and $\nabla_{\theta} L_t(\cdot)$ denotes the gradient of the loss function w.r.t. model parameters $\theta$. The final measure of influence is:
\begin{equation}
\mathcal{I}_\text{TracIn-CP}(z_i, z) = \sum_{t \in \obsteps_\text{CP}} \eta_t \nabla_{\theta} L_{t}(z_i)^\top \nabla_{\theta} L_{t}(z)
\end{equation}

In words, this approximation computes what the loss reduction \emph{would have been} at each $t \in \obsteps_\text{CP}$ under 1) the \emph{simplifying assumption} that the loss function is locally linear in a neighborhood around $\theta_t$, and 2) \emph{if} we had taken a gradient step on example $z_i$ at step $t$ (note that $t$ is an arbitrary checkpoint, so the actual training example encountered at step $t$ was probably not $z_i$). We therefore call this approximation the \emph{hypothetical loss reduction}, to contrast it with the \emph{actual loss reduction}.

We offer a derivation of Equation~\ref{eq:tracincp_approx} to support this interpretation. First, we write the loss as $L_t(z) = L(z, \theta_t)$ to explicitly acknowledge the loss's dependence on $\theta_t$, the model parameters at step $t$. Next, we approximate $L(z, \theta)$ as a linear function of $\theta$ in the neighborhood of $\theta_t$, using the standard first-order Taylor series approximation:
\begin{equation}\label{eq:taylor_approx}
L(z, \theta) \approx L(z, \theta_t) + (\theta - \theta_t)^\top \nabla L(z, \theta_t)
\end{equation}
Then, we use Equation~\ref{eq:taylor_approx} to approximate $L(z, \theta_{t+1})$, the loss at time $t+1$. If we had hypothetically taken a gradient step on example $z_i$ at time $t$, then $\theta_{t+1} = \theta_t -\eta_t \nabla_\theta L(z_i, \theta_t)$. Plugging this into Equation~\ref{eq:taylor_approx}, we get:
\begin{align*}
L(z, \theta_{t+1}) &= L(z, \theta_t - \eta_t \cdot \nabla_\theta L(z_i, \theta_t)) \\
&\approx L(z, \theta_t) - \eta_t \cdot \nabla_\theta L(z_i, \theta_t)^\top \nabla_\theta L(z, \theta_t) \\
&\overset{\text{def}}{=} \tilde{L}_{t+1}(z)
\end{align*}
We call this approximation the \emph{hypothetical loss} at time $t+1$, and denote it $\tilde{L}_{t+1}(z)$ (note the tilde).

Returning to the original formula for \tracinideal{} in Equation~\ref{eq:tracinideal}, we simply replace the true loss $L_t(z)$ with the hypothetical loss $\tilde{L}_t(z)$, and we see that the result is \tracincp{}:
\begin{align*}
L_{t-1}(z) - L_{t}(z) &\approx L_{t-1}(z) - \tilde{L}_{t}(z) \\
&= L_{t-1}(z) - \left(L_{t-1}(z) - \eta_{t-1} \nabla_{\theta} L_{t-1}(z_i)^\top \nabla_{\theta} L_{t-1}(z) \right) \\
&= \eta_{t-1} \nabla_{\theta} L_{t-1}(z_i)^\top \nabla_{\theta} L_{t-1}(z)
\end{align*}

\textbf{In conclusion, \tracincp{} is the same as \tracinideal{}, but where the loss $L_t(z)$ has been replaced by the hypothetical loss $\tilde{L}_t(z)$}.

In isolation, the hypothetical loss is not actually cheaper to compute than the actual loss. However, it has one key advantage. The actual loss $L_t(z)$ must be recorded \emph{while training is happening}, at every step $t$ where training example $z_i$ is actually encountered. In contrast, the hypothetical loss only requires us to save model checkpoints at regular intervals. Then, at any later time, we can use each checkpoint to compute the hypothetical loss for any test example $z$, after a training step on any example $z_i$. In Appendix~\ref{sec:compute}, we provide a deeper analysis of the difference in computational cost between computing actual and hypothetical losses.

Hypothetical losses can also be incorporated into \simflinear{}. Starting with the \simflinear{} objective in Equation~\ref{eq:objective}, we can replace the actual loss $L_t(z)$ with $\tilde{L}_t(z)$, and also only sum over time steps where checkpoints were saved:
\begin{equation}\label{eq:hypo_objective}
\mathcal{O}_\text{hypo}(A, B) = \sum_{t \in \obsteps_\text{CP}} \left(\tilde{L}_t(z) - \hatloss_t(z)\right)^2 + \lambda (\|A\|^2 + \|B\|^2)
\end{equation}
We call this Hypothetical \simflinear{}, and it shares the same advantages as \tracincp{}.

We can apply the same change to our ablation, \simfadd{}. Recall the closed-form solution to \simfadd{} in Equation~\ref{eq:simfadd_soln}. If we replace true losses with hypothetical losses and only sum over saved checkpoints, it becomes:
\begin{equation}
\hat{B}_{i} = -\frac{1}{\obsteps_\text{CP}} \sum_{t \in \obsteps_\text{CP}} L_{t}(z) - \tilde{L}_{t+1}(z)
= -\frac{1}{\obsteps_\text{CP}} \sum_{t \in \obsteps_\text{CP}} \eta_t \nabla_{\theta} L_{t}(z_i)^\top \nabla_{\theta} L_{t}(z)
\end{equation}
Note that $\hat{B}_i$ matches $\mathcal{I}_\text{TracIn-CP}(z, z_i)$ up to a normalization constant. \textbf{Hence, \tracincp{} is equivalent to Hypothetical \simfadd{}.} For this reason, \tracincp{} can be viewed as a purely additive simulator with parameters $B_i$ as defined above.

\subsection{Influence functions}

Influence functions were first developed in the context of robust statistics \citep{hampel1974influence, cook1980characterizations} and later adapted to deep learning by \cite{Koh}. They model the influence of example $z_i$ on example $z$ as:
\begin{equation*}
\mathcal{I}_\text{inf-fns}(z, z_i) = \nabla_\theta L(z_i, \theta_T)^\top H^{-1}_{\theta_T} \nabla_\theta L(z, \theta_T)
\end{equation*}
where $H_{\theta_T}$ is the Hessian of the training loss at the final checkpoint, $\theta_T$.

To draw a connection between Simfluence and influence functions, we continue to build on the notion of \emph{hypothetical training steps} from the previous section. Let $\theta_T$ be the final model checkpoint. Then, imagine a hypothetical training step on example $z_i$. For first-order gradient descent, the new parameters would be $\theta_{T+1} = \theta_T - \eta_T \nabla_\theta L(z_i, \theta_T)$. But for second-order gradient descent (e.g. Newton's method), the parameters would be:
\begin{equation*}
\theta_{T+1} = \theta_T - H^{-1}_{\theta_T} \nabla_\theta L(z_i, \theta_T)
\end{equation*}
If we plug this into the Taylor series approximation from Equation~\ref{eq:taylor_approx}, then the hypothetical loss at time $T+1$ would be:
\begin{equation}
\tilde{L}_{T+1}(z) = L(z, \theta_T) - \nabla_\theta L(z_i, \theta_T)^\top H^{-1}_{\theta_T} \nabla_\theta L(z, \theta_T)
\end{equation}
Similar to the previous section, we can replace true observed losses with second-order hypothetical losses in our objective function for \simfadd{}. Then, the closed form solution (following Equation~\ref{eq:simfadd_soln}) is:
\begin{equation*}
\hat{B}_i = -\nabla_\theta L(z_i, \theta_T)^\top H^{-1}_{\theta_T} \nabla_\theta L(z, \theta_T)
\end{equation*}
Each $-\hat{B}_i$ is exactly equal to the influence score defined by influence functions, $\mathcal{I}_\text{inf-fns}(z, z_i)$. Hence, \textbf{Influence functions are equivalent to Hypothetical \simfadd{} when using second-order hypothetical losses.} As such, Hessian-based influence also models only the additive terms in our simulation.

For our evaluations, we focus on comparisons to first-order methods (TracIn), as these have been shown to scale better with model and dataset size~\citep{yeh2018representer,Koh,sogaard2021revisiting} and the integration over timesteps aligns more closely with our time-series simulation paradigm.

\section{Experiments}\label{sec:experiments}

Now that we have presented our proposed approach (\simflinear{}) and prior TDA methods as training run simulators, we can evaluate and compare their simulation accuracy using the metrics defined in Section~\ref{sec:eval-metrics}: all-steps MSE (which measures a simulator's accuracy in predicting an example's loss at each step of training) and final-step Spearman's $\rho$ (which measures a simulator's ability to predict the relative ordering of losses among test examples at the end of training). Our experiments simulate large language model (LLM) fine-tuning on different datasets and training methods, described below.

\paragraph{LLM fine-tuning methods and models.}
We consider two different LLM fine-tuning methods: standard full-model tuning where all model parameters are updated, and parameter-efficient tuning where only a subset of model parameters are updated. For standard full-model tuning, we use either T0 3B \citep{sanh2022multitask} or T5 LM-Adapted XL \citep[T5-LMA;][]{lester-etal-2021-power}. For few-shot parameter-efficient tuning, we apply the IA3 method of \citet {liu2022tfew} to T0 3B. We use a batch size of 4 for all training runs.

\paragraph{Datasets.}
We perform LLM fine-tuning on three datasets: Recognizing Textual Entailments \citep[RTE;][]{dagan2006pascal}, Choice of Plausible Alternatives \citep[COPA;][]{roemmele2011choice}, and Winogrande \citep{sakaguchi2021winogrande}. These datasets are well-studied in the literature, and all tasks of their categories (natural language inference, next sentence prediction, and coreference resolution) are specifically held out from the instruction tuning mixture of T0. For RTE, the model must generate either ``Yes'' or ``No''. For COPA and Winogrande, the model is presented with multiple free-text options, and must generate the correct option (which sentence should be the next sentence or which referent does a pronoun refer to) --- unlike RTE, these options do not come from a fixed set of labels.

\paragraph{Training runs.}
For RTE and COPA, we consider a few-shot setting. For each run, we train on 64 examples randomly selected from a fixed pool of 100 examples, such that each training run involves a different set of 64 examples. Each run is 4 epochs, and examples are randomly shuffled for each epoch. For Winogrande, we study a non-few-shot setting. For each run, we train on 1024 examples randomly selected from a fixed pool of 1536 examples. Each run is 1.5 epochs, with examples randomly shuffled for each epoch.

For each dataset and fine-tuning method, we perform 32 training runs (as described above). We then randomly split these training runs into $\rpastset$ (22 runs) and $\rfutureset$ (10 runs). We use $\rpastset$ for training our simulators: 20 runs for fitting simulator parameters $\Omega = (A, B)$, and 2 runs as a validation set for finding the best setting of $\lambda$, the L2-regularization hyperparameter. We use $\rfutureset$ as held-out runs for evaluation: we report simulation accuracy metrics on this set in Tables~\ref{tab:tracin} and~\ref{tab:main-result}.

\paragraph{Adjustments for \tracincp{}.}

In Section~\ref{sec:connections}, we noted that \tracincp{} uses the approximation $L_{t}(z) - L_{t+1}(z) \approx\eta_t \nabla_{\theta} L_{t}(z_i)^\top \nabla_{\theta} L_{t}(z)$. This approximation assumes that each training step applies vanilla gradient descent with learning rate $\eta_t$ --- this assumption is actually false in most LLM fine-tuning setups, where training steps typically use Adam \citep{kingma2014adam} or Adafactor \citep{shazeer2018adafactor}, which produce parameter updates that have significantly different magnitude than vanilla gradient descent.

For this reason, we empirically observed that the loss trajectories predicted by \tracincp{} would often have a reasonable ``shape'', but the wrong scale. This did not affect our Spearman's $\rho$ results (which are only sensitive to relative orderings), but did result in very high MSE. To strengthen \tracincp{}, we rescale its predicted loss trajectories by an optimal factor $\sigma$, chosen to minimize $\sum_{t=1}^T (\sigma \hatloss_t(z) - L_t(z))^2$. Note that this gives \tracincp{} an unfair advantage over other methods, since the scaling factor depends on the ground truth losses, $L_{1:T}$. Our experiments show that \simflinear{} still outperforms \tracincp{}, even when it is optimally rescaled.

Another issue with \tracincp{} stems from its memory requirements. \tracincp{} requires the user to compute and store model gradients for every example at every checkpoint. For LLMs with billions of parameters, saving full model gradients for every training example requires a prohibitive amount of storage. Hence, most applications of \tracincp{} resort to approximations such as only saving gradients for one particular layer of the LLM. To avoid error introduced by such approximations, we only evaluate \tracincp{} for parameter-efficient tuning. In this setting, only a small number of parameters are being updated, so the gradients to store are orders of magnitude smaller. In contrast, Simfluence does not depend on model checkpoints or anything that scales with model size, so it is equally applicable to all fine-tuning methods.

\begin{table}[t]
\centering
\renewcommand{\arraystretch}{1.2}
\begin{tabular}{clll} 
\toprule
\multicolumn{1}{l}{Task} & TDA Method                             & \begin{tabular}[c]{@{}l@{}}All-steps Mean \\Squared Error\end{tabular} & \begin{tabular}[c]{@{}l@{}}Final-step \\Spearman's $\rho$\end{tabular}  \\ 
\midrule
\multirow{5}{*}{COPA}       & \tracincp{} (10 ckpts)                  & $5.873 _{\pm 0.307}$                                                      & $0.448 _{\pm 0.079}$                                                       \\
                            & \tracincp{} (all steps)                 & $5.792 _{\pm 0.331}$                                                      & $0.469 _{\pm 0.077}$                                                       \\
                            & \simfadd{} (\tracinideal{}) & $2.506 _{\pm 0.491}$                                                      & $0.786 _{\pm 0.065}$                                                       \\
                            & \simfmult{}          & $1.557 _{\pm 0.469}$                                                      & $0.763 _{\pm 0.046}$                                                       \\
                            & \simflinear{}                  & $\mathbf{1.503 _{\pm 0.494}}$                                             & $\mathbf{0.886 _{\pm 0.033}}$                                              \\ 
\hline
\multirow{5}{*}{RTE}        & \tracincp{} (10 ckpts)                  & $3.317 _{\pm 0.323}$                                                      & $0.040 _{\pm 0.202}$                                                       \\
                            & \tracincp{} (all steps)                 & $3.733 _{\pm 0.275}$                                                      & $0.034 _{\pm 0.200}$                                                       \\
                            & \simfadd{} (\tracinideal{}) & $3.096 _{\pm 3.161}$                                                      & $0.543 _{\pm 0.288}$                                                       \\
                            & \simfmult{}          & $5.818 _{\pm 14.311}$                                                     & $0.599 _{\pm 0.182}$                                                       \\
                            & \simflinear{}                  & $\mathbf{0.819 _{\pm 1.222}}$                                             & $\mathbf{0.887 _{\pm 0.080}}$                                              \\
\bottomrule
\end{tabular}
\caption{Comparison between Simfluence and \tracincp{}. Results are averaged over 10 held-out test runs. Lower MSE is better, and higher Spearman's $\rho$ is better. See \autoref{sec:eval-metrics} for the full definitions of the metrics.}  
\label{tab:tracin}
\renewcommand{\arraystretch}{1}
\end{table}

\begin{table}[t]
\centering
\renewcommand{\arraystretch}{1.2}
\begin{tabular}{ccclll} 
\toprule
\multicolumn{1}{l}{Task}    & \multicolumn{1}{l}{\begin{tabular}[c]{@{}l@{}}Fine-tuning\\Method\end{tabular}} & \multicolumn{1}{l}{\begin{tabular}[c]{@{}l@{}}Language\\Model\end{tabular}} & \begin{tabular}[c]{@{}l@{}}Simfluence\\Model\end{tabular} & \begin{tabular}[c]{@{}l@{}}All Steps Mean\\ Squared Error\end{tabular} & \begin{tabular}[c]{@{}l@{}}Final Step \\Spearman's $\rho$\end{tabular}  \\ 
\midrule
\multirow{9}{*}{COPA}       & \multirow{6}{*}{FMT}                                                            & \multirow{3}{*}{T0}                                                         & \textsc{additive}                                                  & $0.296 _{\pm 0.163}$                                                      & $0.278 _{\pm 0.141}$                                                       \\
                            &                                                                                 &                                                                             & \textsc{multiplicative}                                            & $1.582 _{\pm 4.274}$                                                      & $\mathbf{0.629 _{\pm 0.058}}$                                              \\
                            &                                                                                 &                                                                             & \textsc{linear}                                                    & $\mathbf{0.272 _{\pm 0.425}}$                                             & $\mathbf{0.628 _{\pm 0.065}}$                                              \\ 
\cline{3-6}
                            &                                                                                 & \multirow{3}{*}{T5-LMA}                                                     & \textsc{additive}                                                  & $0.801 _{\pm 0.216}$                                                      & $0.368 _{\pm 0.117}$                                                       \\
                            &                                                                                 &                                                                             & \textsc{multiplicative}                                            & $0.240 _{\pm 0.158}$                                                      & $\mathbf{0.672 _{\pm 0.079}}$                                              \\
                            &                                                                                 &                                                                             & \textsc{linear}                                                    & $\mathbf{0.175 _{\pm 0.113}}$                                             & $0.565 _{\pm 0.085}$                                                       \\ 
\cline{2-6}
                            & \multirow{3}{*}{IA3}                                                            & \multirow{3}{*}{T0}                                                         & \textsc{additive}                                                  & $2.506 _{\pm 0.491}$                                                      & $0.786 _{\pm 0.065}$                                                       \\
                            &                                                                                 &                                                                             & \textsc{multiplicative}                                            & $1.557 _{\pm 0.469}$                                                      & $0.763 _{\pm 0.046}$                                                       \\
                            &                                                                                 &                                                                             & \textsc{linear}                                                    & $\mathbf{1.503 _{\pm 0.494}}$                                             & $\mathbf{0.886 _{\pm 0.033}}$                                              \\ 
\hline
\multirow{9}{*}{RTE}        & \multirow{6}{*}{FMT}                                                            & \multirow{3}{*}{T0}                                                         & \textsc{additive}                                                  & $7.385 _{\pm 7.696}$                                                      & $0.451 _{\pm 0.374}$                                                       \\
                            &                                                                                 &                                                                             & \textsc{multiplicative}                                            & $6.818 _{\pm 5.063}$                                                      & $0.609 _{\pm 0.233}$                                                       \\
                            &                                                                                 &                                                                             & \textsc{linear}                                                    & $\mathbf{6.709 _{\pm 14.756}}$                                            & $\mathbf{0.813 _{\pm 0.116}}$                                              \\ 
\cline{3-6}
                            &                                                                                 & \multirow{3}{*}{T5-LMA}                                                     & \textsc{additive}                                                  & $14.531 _{\pm 8.214}$                                                     & $0.122 _{\pm 0.193}$                                                       \\
                            &                                                                                 &                                                                             & \textsc{multiplicative}                                            & $2.866 _{\pm 3.141}$                                                      & $0.086 _{\pm 0.087}$                                                       \\
                            &                                                                                 &                                                                             & \textsc{linear}                                                    & $\mathbf{2.290 _{\pm 2.575}}$                                             & $\mathbf{0.399 _{\pm 0.266}}$                                              \\ 
\cline{2-6}
                            & \multirow{3}{*}{IA3}                                                            & \multirow{3}{*}{T0}                                                         & \textsc{additive}                                                  & $3.096 _{\pm 3.161}$                                                      & $0.543 _{\pm 0.288}$                                                       \\
                            &                                                                                 &                                                                             & \textsc{multiplicative}                                            & $5.818 _{\pm 14.311}$                                                     & $0.599 _{\pm 0.182}$                                                       \\
                            &                                                                                 &                                                                             & \textsc{linear}                                                    & $\mathbf{0.819 _{\pm 1.222}}$                                             & $\mathbf{0.887 _{\pm 0.080}}$                                              \\ 
\hline
\multirow{3}{*}{Winogrande} & \multirow{3}{*}{FMT}                                                            & \multirow{3}{*}{T5-LMA}                                                     & \textsc{additive}                                                  & $7.060 _{\pm 1.961}$                                                      & $0.256 _{\pm 0.156}$                                                       \\
                            &                                                                                 &                                                                             & \textsc{multiplicative}                                            & $1.496 _{\pm 0.400}$                                                      & $0.339 _{\pm 0.190}$                                                       \\
                            &                                                                                 &                                                                             & \textsc{linear}                                                    & $\mathbf{0.910 _{\pm 0.104}}$                                             & $\mathbf{0.383 _{\pm 0.151}}$                                              \\
\bottomrule
\end{tabular}
\renewcommand{\arraystretch}{1}  %
\caption{Quality of fit between our simulator's predicted losses and the ground-truth losses of 10 held-out runs.
Standard deviation is reported after the $\pm$ sign.
\simfadd{} is equivalent to \tracinideal{} (\citealp{Pruthi}; see \autoref{sec:connections} for a proof). 
FMT = full-model tuning. IA3 = parameter-efficient tuning \citet{liu2022tfew}.}
\label{tab:main-result}
\end{table}

\paragraph{Results.}

\autoref{tab:tracin} shows our primary results: a comparison of our proposed method (\simflinear{}) and its ablations (\simfadd{} and \simfmult{}) to existing methods (\tracincp{} and \tracinideal{}), evaluated on how well their simulated loss trajectories match the loss trajectories of real training runs. For these results, we used IA3 parameter-efficient tuning on T0 3B.

Following the typical usage of \tracincp{}, we select 10 checkpoints for \tracincp{} over which to accumulate losses. However, as noted in Section~\ref{sec:tracin-cp}, accumulating losses over just 10 checkpoints (rather than all steps) is an approximation that may introduce error. Therefore, we also evaluate \tracincp{} in its best-case scenario where we accumulate losses over all training steps of all 20 training runs.

\autoref{tab:tracin} yields several important insights. First, all Simfluence variants outperform all \tracin{} variants. They have smaller MSE, meaning they predict loss trajectories better, and they have much higher Spearman's $\rho$, meaning they predict the final loss of each example better. As noted in Section~\ref{sec:connections}, the only difference between \simfadd{} and \tracincp{} (all steps) is that the former uses \emph{actual observed losses} while the latter uses \emph{hypothetical losses} estimated from saved model checkpoints. Both are purely additive models of influence. Therefore, the significant gap in quality between these two approaches can be attributed to the approximation error introduced by hypothetical losses.

\begin{figure}[t]
     \centering
     \begin{subfigure}[b]{0.49\textwidth}
         \centering
         \includegraphics[width=\textwidth]{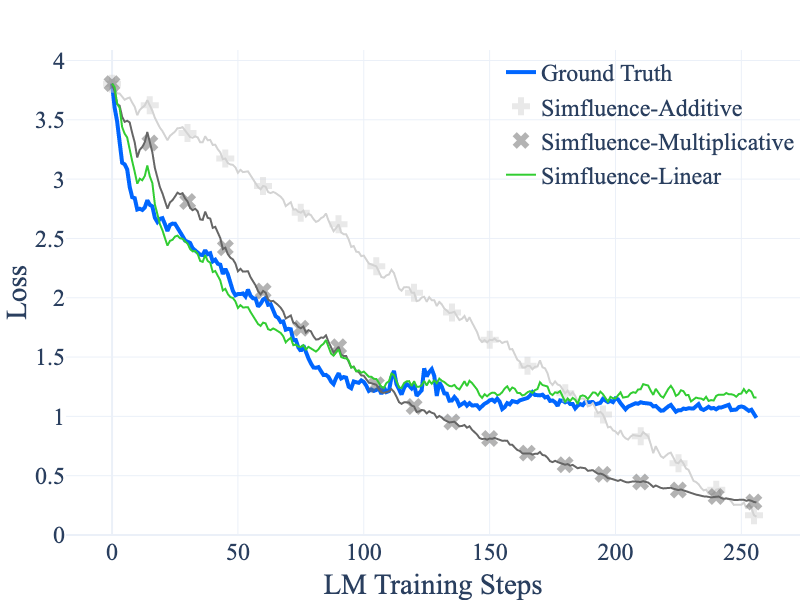}
         \caption{COPA} %
         \label{fig:copa-cherry}
     \end{subfigure}
     \hfill
     \begin{subfigure}[b]{0.49\textwidth}
         \centering
         \includegraphics[width=\textwidth]{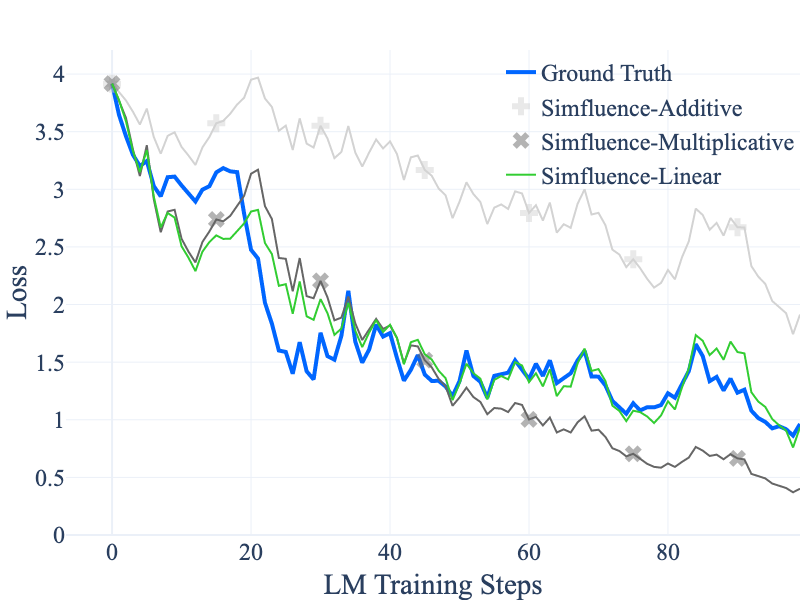}
         \caption{RTE} %
         \label{fig:rte-cherry}
     \end{subfigure}
     \caption{Qualitative examples of Simfluence's predicted loss trajectories on the loss of one random test example in one run.}
     \label{fig:cherries}
\end{figure}

Second, we see that \simflinear{} strongly outperforms both \simfadd{} (equivalent to \tracinideal{}) and \simfmult{}. This shows that it is important to model \emph{both additive and multiplicative} influence. This improvement is also qualitatively visible: 
\autoref{fig:cherries} shows that \simflinear{} (green lines) predicts the shape of true held-out loss trajectories (blue lines) better than either \simfadd{} or \simfmult{} (gray lines with plus/multiply signs).

\autoref{tab:main-result} continues to evaluate Simfluence on a wider range of tasks (RTE, COPA, Winogrande), language models (T0, T5), fine-tuning methods (standard full-model tuning, parameter-efficient tuning), and numbers of examples (few-shot, many-shot). Again, \simflinear{} gives significant gains over both \simfmult{} and \simfadd{} in nearly all settings, reinforcing the importance of modeling both additive and multiplicative influence. Interestingly, when we compare the two ablations (\simfadd{} and \simfmult{}), neither is strictly better than the other across all setups, suggesting that influence may be more additive or more multiplicative depending on the task or dataset.

Next, we restrict our attention to just \simflinear{}. We find that simulation accuracy varies across setups. If we focus on all-steps mean squared error, there does not appear to be a clear trend. On the other hand, if we focus on final-step Spearman's $\rho$, we find that \simflinear{} appears to best at simulating T0 IA3 fine-tuning (roughly 0.88 Spearman on both COPA and RTE), does a little worse on T0 full-model tuning (Spearman ranges between 0.628 and 0.813), and the least well on T5-LMA full-model tuning (Spearman ranges between 0.383 and 0.565). Further research is needed to study and explain these differences.

\section{Related work}\label{sec:related}

As described in Section~\ref{sec:connections}, our method is closely related to other TDA methods for estimating the influence of training examples on test predictions. Prior methods use gradients for tractability; in this work, we explore settings where that approximation is not needed and where we can directly measure loss deltas from repeated training runs. Because our method is based on the full loss trajectory over the course of training, it bears the closest relationship to \tracin{} \citep{Pruthi}, in contrast to \citet{Koh}, \citet{schioppa2022scaling}, and \citet{guo2021fastif} which use Hessian-based approximations on the final trained model.
Recently, these methods have been applied to explain predictions and identify data artifacts in a number of NLP tasks, including NLI \citep{koh2019accuracy, han2020explaining, pezeshkpour2022combining} and language model pretraining \citep{han2022orca}.

Very closely related to our work is \citet{sogaard2021revisiting}, which observes that existing influence methods are a poor approximation of leave-one-out accuracy and do not account for training data order, to which they are highly sensitive. Additionally, \citet{koh2019accuracy} study influence in the group setting where multiple training points are left out, and find high correlation between predicted and actual effects, although the actual errors are large and the predicted effects tend to be an underestimate. \citet{han-tsvetkov-2021-influence-tuning} evaluate using influence of specific examples as a training objective, in order to guide the model away from reliance on spurious correlations. In work contemporaneous to ours, \citet{yang2023remove} also use influence functions to guide targeted interventions, using them as a heuristic to identify training examples to remove in order to flip a target prediction, although they only evaluate with a convex model (logistic regression).

While the above training data attribution methods start from an interpretability perspective, Simfluence can also be viewed as a simulation of the training process, and in this respect is related to more general work on learning dynamics of deep networks. Specifically, Simfluence bears resemblence to prior work on curriculum design; see \citet{wang2021survey} for an overview. For example, \cite{kim2018screenernet,fan2017learning} use deep reinforcement learning in order to select an optimal training curriculum, and \citet{jiang2018mentornet} train an auxiliary neural network model in order to improve the primary model's generalization performance. Similar to our approach, \cite{swayamdipta-etal-2020-dataset} use a model's loss on and confidence in individual training examples in order to ``map'' the training set as a whole, and provide insights regarding which examples best support e.g., out of distribution generalization. In general, prior work on curricula focus on characterizing ``easy'' and ``hard'' examples, with the goal of improving overall model performance. In contrast, our focus is on interpretability and training data attribution for specific predictions at test time.

\section{Limitations and future work}\label{sec:limitations}

\paragraph{Parameter-efficiency.}
In Section~\ref{sec:approach}, we noted that \simflinear{} is not particularly data-efficient: each simulator must learn two parameters ($A_i$ and $B_i$) for every training example of interest, which ultimately requires us to observe at least $2n$ training steps if we wish to simulate $n$ training examples. In future work, we hope to explore more data-efficient, featurized simulators --- for example, instead of learning a separate simulator for every test example, and learning separate parameters $(A_{i}, B_{i})$ for every training example, we could model $\alpha(c_t)$ and $\beta(c_t)$ as:
\begin{equation*}
\alpha(c_t) = \sum_{i \in c_t} \Phi(z_i)^\top \Phi(z) \qquad \beta(c_t) = \sum_{i \in c_t} \Psi(z_i)^\top \Psi(z)
\end{equation*}
where $\Phi(z)$ and $\Psi(z)$ are neural network encoders that map any example to a low-dimensional vector.\footnote{Like \tracincp{}, we are now modeling influence as a dot-product of low-dimensional vectors, but the vectors are no longer loss gradients.} More generally, any kind of learned model can be used to parameterize both $\alpha(c_t)$ and $\beta(c_t)$. This may increase the generalization of the simulator, as well as its sample efficiency.

\paragraph{Handling previously unseen examples.}
The above proposal also overcomes another current limitation of \simflinear{}: it cannot simulate training runs that include examples which have never been seen before in $\rpastset$. This problem is eliminated if we switch to a featurized representation of examples, proposed above.

\paragraph{Expressive power.}
We note that while \simflinear{} can model redundancy, it still cannot model other \emph{supermodular interactions} between examples. For example, if example $z_1$ teaches a model that ``Paris is in France'', and a later example $z_2$ teaches a model that ``France is in Europe'', then the two examples combined may be enough for a model to learn that ``Paris is in Europe'', while neither alone would be sufficient. For this, we would need to move beyond a simple Markov model to something that allows for longer-range interactions between examples --- a direction we leave for future work.

There is also the more general point that \simflinear{} oversimplifies the real dynamics of training. In reality, the loss at each time step is governed by the model's underlying parameters, optimizer variables and learning rate --- aspects that we have ignored in \simflinear{}.

While these are limitations of \simflinear{}, they are not limitations of the general \emph{training run simulation} paradigm (Simfluence): future work could explore a wide range of time series models to more accurately simulate training. The only requirements are that the simulator should run much faster than actual training, and should be learnable from a modest number of previously observed runs.

\subsubsection*{Acknowledgments}

We would like to thank Daphne Ippolito, Deepak Ramachandran, Kathy Meier-Hellstern, Arun Chaganty and Raphael Hoffmann for their insightful feedback on the paper.

\bibliography{main}

\begin{thebibliography}{35}
\providecommand{\natexlab}[1]{#1}
\providecommand{\url}[1]{\texttt{#1}}
\expandafter\ifx\csname urlstyle\endcsname\relax
  \providecommand{\doi}[1]{doi: #1}\else
  \providecommand{\doi}{doi: \begingroup \urlstyle{rm}\Url}\fi

\bibitem[Bengio et~al.(2009)Bengio, Louradour, Collobert, and
  Weston]{bengio2009curriculum}
Yoshua Bengio, J{\'e}r{\^o}me Louradour, Ronan Collobert, and Jason Weston.
\newblock Curriculum learning.
\newblock In \emph{Proceedings of the 26th annual international conference on
  machine learning}, pp.\  41--48, 2009.

\bibitem[Brown et~al.(2020)Brown, Mann, Ryder, Subbiah, Kaplan, Dhariwal,
  Neelakantan, Shyam, Sastry, Askell, et~al.]{brown2020language}
Tom Brown, Benjamin Mann, Nick Ryder, Melanie Subbiah, Jared~D Kaplan, Prafulla
  Dhariwal, Arvind Neelakantan, Pranav Shyam, Girish Sastry, Amanda Askell,
  et~al.
\newblock Language models are few-shot learners.
\newblock \emph{Advances in neural information processing systems},
  33:\penalty0 1877--1901, 2020.

\bibitem[Cook \& Weisberg(1980)Cook and Weisberg]{cook1980characterizations}
R~Dennis Cook and Sanford Weisberg.
\newblock Characterizations of an empirical influence function for detecting
  influential cases in regression.
\newblock \emph{Technometrics}, 22\penalty0 (4):\penalty0 495--508, 1980.

\bibitem[Dagan et~al.(2006)Dagan, Glickman, and Magnini]{dagan2006pascal}
Ido Dagan, Oren Glickman, and Bernardo Magnini.
\newblock The pascal recognising textual entailment challenge.
\newblock In \emph{Machine Learning Challenges. Evaluating Predictive
  Uncertainty, Visual Object Classification, and Recognising Tectual
  Entailment: First PASCAL Machine Learning Challenges Workshop, MLCW 2005,
  Southampton, UK, April 11-13, 2005, Revised Selected Papers}, pp.\  177--190.
  Springer, 2006.

\bibitem[Deng et~al.(2009)Deng, Dong, Socher, Li, Li, and
  Fei-Fei]{deng2009imagenet}
Jia Deng, Wei Dong, Richard Socher, Li-Jia Li, Kai Li, and Li~Fei-Fei.
\newblock Imagenet: A large-scale hierarchical image database.
\newblock In \emph{2009 IEEE conference on computer vision and pattern
  recognition}, pp.\  248--255. Ieee, 2009.

\bibitem[Fan et~al.(2017)Fan, Tian, Qin, Bian, and Liu]{fan2017learning}
Yang Fan, Fei Tian, Tao Qin, Jiang Bian, and Tie-Yan Liu.
\newblock Learning what data to learn.
\newblock \emph{arXiv preprint arXiv:1702.08635}, 2017.

\bibitem[Fujishige(2005)]{fujishige2005submodular}
Satoru Fujishige.
\newblock \emph{Submodular functions and optimization}.
\newblock Elsevier, 2005.

\bibitem[Guo et~al.(2021)Guo, Rajani, Hase, Bansal, and Xiong]{guo2021fastif}
Han Guo, Nazneen Rajani, Peter Hase, Mohit Bansal, and Caiming Xiong.
\newblock Fastif: Scalable influence functions for efficient model
  interpretation and debugging.
\newblock In \emph{Proceedings of the 2021 Conference on Empirical Methods in
  Natural Language Processing}, pp.\  10333--10350, 2021.

\bibitem[Halevy et~al.(2009)Halevy, Norvig, and
  Pereira]{halevy2009unreasonable}
Alon Halevy, Peter Norvig, and Fernando Pereira.
\newblock The unreasonable effectiveness of data.
\newblock \emph{IEEE intelligent systems}, 24\penalty0 (2):\penalty0 8--12,
  2009.

\bibitem[Hampel(1974)]{hampel1974influence}
Frank~R Hampel.
\newblock The influence curve and its role in robust estimation.
\newblock \emph{Journal of the american statistical association}, 69\penalty0
  (346):\penalty0 383--393, 1974.

\bibitem[Han \& Tsvetkov(2021)Han and
  Tsvetkov]{han-tsvetkov-2021-influence-tuning}
Xiaochuang Han and Yulia Tsvetkov.
\newblock Influence tuning: Demoting spurious correlations via instance
  attribution and instance-driven updates.
\newblock In \emph{Findings of the Association for Computational Linguistics:
  EMNLP 2021}, pp.\  4398--4409, Punta Cana, Dominican Republic, November 2021.
  Association for Computational Linguistics.
\newblock \doi{10.18653/v1/2021.findings-emnlp.374}.
\newblock URL \url{https://aclanthology.org/2021.findings-emnlp.374}.

\bibitem[Han \& Tsvetkov(2022)Han and Tsvetkov]{han2022orca}
Xiaochuang Han and Yulia Tsvetkov.
\newblock Orca: Interpreting prompted language models via locating supporting
  data evidence in the ocean of pretraining data.
\newblock \emph{arXiv preprint arXiv:2205.12600}, 2022.

\bibitem[Han et~al.(2020)Han, Wallace, and Tsvetkov]{han2020explaining}
Xiaochuang Han, Byron~C Wallace, and Yulia Tsvetkov.
\newblock Explaining black box predictions and unveiling data artifacts through
  influence functions.
\newblock In \emph{Proceedings of the 58th Annual Meeting of the Association
  for Computational Linguistics}, pp.\  5553--5563, 2020.

\bibitem[Hoerl \& Kennard(1970)Hoerl and Kennard]{hoerl1970ridge}
Arthur~E Hoerl and Robert~W Kennard.
\newblock Ridge regression: Biased estimation for nonorthogonal problems.
\newblock \emph{Technometrics}, 12\penalty0 (1):\penalty0 55--67, 1970.

\bibitem[Jiang et~al.(2018)Jiang, Zhou, Leung, Li, and
  Fei-Fei]{jiang2018mentornet}
Lu~Jiang, Zhengyuan Zhou, Thomas Leung, Li-Jia Li, and Li~Fei-Fei.
\newblock Mentornet: Learning data-driven curriculum for very deep neural
  networks on corrupted labels.
\newblock In \emph{International conference on machine learning}, pp.\
  2304--2313. PMLR, 2018.

\bibitem[Kaplan et~al.(2020)Kaplan, McCandlish, Henighan, Brown, Chess, Child,
  Gray, Radford, Wu, and Amodei]{kaplan2020scaling}
Jared Kaplan, Sam McCandlish, Tom Henighan, Tom~B Brown, Benjamin Chess, Rewon
  Child, Scott Gray, Alec Radford, Jeffrey Wu, and Dario Amodei.
\newblock Scaling laws for neural language models.
\newblock \emph{arXiv preprint arXiv:2001.08361}, 2020.

\bibitem[Kim \& Choi(2018)Kim and Choi]{kim2018screenernet}
Tae-Hoon Kim and Jonghyun Choi.
\newblock Screenernet: Learning self-paced curriculum for deep neural networks.
\newblock \emph{arXiv preprint arXiv:1801.00904}, 2018.

\bibitem[Kingma \& Ba(2014)Kingma and Ba]{kingma2014adam}
Diederik~P Kingma and Jimmy Ba.
\newblock Adam: A method for stochastic optimization.
\newblock \emph{arXiv preprint arXiv:1412.6980}, 2014.

\bibitem[Koh \& Liang(2017)Koh and Liang]{Koh}
Pang~Wei Koh and Percy Liang.
\newblock Understanding black-box predictions via influence functions.
\newblock In \emph{International conference on machine learning}, pp.\
  1885--1894. PMLR, 2017.

\bibitem[Koh et~al.(2019)Koh, Ang, Teo, and Liang]{koh2019accuracy}
Pang Wei~W Koh, Kai-Siang Ang, Hubert Teo, and Percy~S Liang.
\newblock On the accuracy of influence functions for measuring group effects.
\newblock \emph{Advances in neural information processing systems}, 32, 2019.

\bibitem[Lester et~al.(2021)Lester, Al-Rfou, and
  Constant]{lester-etal-2021-power}
Brian Lester, Rami Al-Rfou, and Noah Constant.
\newblock The power of scale for parameter-efficient prompt tuning.
\newblock In \emph{Proceedings of the 2021 Conference on Empirical Methods in
  Natural Language Processing}, pp.\  3045--3059, Online and Punta Cana,
  Dominican Republic, November 2021. Association for Computational Linguistics.
\newblock \doi{10.18653/v1/2021.emnlp-main.243}.
\newblock URL \url{https://aclanthology.org/2021.emnlp-main.243}.

\bibitem[Liu et~al.(2022)Liu, Tam, Muqeeth, Mohta, Huang, Bansal, and
  Raffel]{liu2022tfew}
Haokun Liu, Derek Tam, Mohammed Muqeeth, Jay Mohta, Tenghao Huang, Mohit
  Bansal, and Colin Raffel.
\newblock Few-shot parameter-efficient fine-tuning is better and cheaper than
  in-context learning.
\newblock \emph{arXiv preprint arXiv:2205.05638}, 2022.

\bibitem[Pezeshkpour et~al.(2022)Pezeshkpour, Jain, Singh, and
  Wallace]{pezeshkpour2022combining}
Pouya Pezeshkpour, Sarthak Jain, Sameer Singh, and Byron~C Wallace.
\newblock Combining feature and instance attribution to detect artifacts.
\newblock In \emph{Findings of the Association for Computational Linguistics:
  ACL 2022}, pp.\  1934--1946, 2022.

\bibitem[Pruthi et~al.(2020)Pruthi, Liu, Sundararajan, and Kale]{Pruthi}
Garima Pruthi, Frederick Liu, Mukund Sundararajan, and Satyen Kale.
\newblock Estimating training data influence by tracking gradient descent.
\newblock \emph{CoRR}, abs/2002.08484, 2020.
\newblock URL \url{https://arxiv.org/abs/2002.08484}.

\bibitem[Roemmele et~al.(2011)Roemmele, Bejan, and Gordon]{roemmele2011choice}
Melissa Roemmele, Cosmin~Adrian Bejan, and Andrew~S Gordon.
\newblock Choice of plausible alternatives: An evaluation of commonsense causal
  reasoning.
\newblock In \emph{AAAI spring symposium: logical formalizations of commonsense
  reasoning}, pp.\  90--95, 2011.

\bibitem[Sakaguchi et~al.(2021)Sakaguchi, Bras, Bhagavatula, and
  Choi]{sakaguchi2021winogrande}
Keisuke Sakaguchi, Ronan~Le Bras, Chandra Bhagavatula, and Yejin Choi.
\newblock Winogrande: An adversarial winograd schema challenge at scale.
\newblock \emph{Communications of the ACM}, 64\penalty0 (9):\penalty0 99--106,
  2021.

\bibitem[Sanh et~al.(2022)Sanh, Webson, Raffel, Bach, Sutawika, Alyafeai,
  Chaffin, Stiegler, Scao, Raja, Dey, Bari, Xu, Thakker, Sharma, Szczechla,
  Kim, Chhablani, Nayak, Datta, Chang, Jiang, Wang, Manica, Shen, Yong, Pandey,
  Bawden, Wang, Neeraj, Rozen, Sharma, Santilli, Fevry, Fries, Teehan, Bers,
  Biderman, Gao, Wolf, and Rush]{sanh2022multitask}
Victor Sanh, Albert Webson, Colin Raffel, Stephen~H. Bach, Lintang Sutawika,
  Zaid Alyafeai, Antoine Chaffin, Arnaud Stiegler, Teven~Le Scao, Arun Raja,
  Manan Dey, M~Saiful Bari, Canwen Xu, Urmish Thakker, Shanya~Sharma Sharma,
  Eliza Szczechla, Taewoon Kim, Gunjan Chhablani, Nihal Nayak, Debajyoti Datta,
  Jonathan Chang, Mike Tian-Jian Jiang, Han Wang, Matteo Manica, Sheng Shen,
  Zheng~Xin Yong, Harshit Pandey, Rachel Bawden, Thomas Wang, Trishala Neeraj,
  Jos Rozen, Abheesht Sharma, Andrea Santilli, Thibault Fevry, Jason~Alan
  Fries, Ryan Teehan, Tali Bers, Stella Biderman, Leo Gao, Thomas Wolf, and
  Alexander~M. Rush.
\newblock Multitask prompted training enables zero-shot task generalization.
\newblock 2022.

\bibitem[Schioppa et~al.(2022)Schioppa, Zablotskaia, Vilar, and
  Sokolov]{schioppa2022scaling}
Andrea Schioppa, Polina Zablotskaia, David Vilar, and Artem Sokolov.
\newblock Scaling up influence functions.
\newblock In \emph{Proceedings of the AAAI Conference on Artificial
  Intelligence}, volume~36, pp.\  8179--8186, 2022.

\bibitem[Shapley et~al.(1953)]{shapley1953value}
Lloyd~S Shapley et~al.
\newblock A value for n-person games.
\newblock 1953.

\bibitem[Shazeer \& Stern(2018)Shazeer and Stern]{shazeer2018adafactor}
Noam Shazeer and Mitchell Stern.
\newblock Adafactor: Adaptive learning rates with sublinear memory cost.
\newblock In \emph{International Conference on Machine Learning}, pp.\
  4596--4604. PMLR, 2018.

\bibitem[S{\o}gaard et~al.(2021)]{sogaard2021revisiting}
Anders S{\o}gaard et~al.
\newblock Revisiting methods for finding influential examples.
\newblock \emph{arXiv preprint arXiv:2111.04683}, 2021.

\bibitem[Swayamdipta et~al.(2020)Swayamdipta, Schwartz, Lourie, Wang,
  Hajishirzi, Smith, and Choi]{swayamdipta-etal-2020-dataset}
Swabha Swayamdipta, Roy Schwartz, Nicholas Lourie, Yizhong Wang, Hannaneh
  Hajishirzi, Noah~A. Smith, and Yejin Choi.
\newblock Dataset cartography: Mapping and diagnosing datasets with training
  dynamics.
\newblock In \emph{Proceedings of the 2020 Conference on Empirical Methods in
  Natural Language Processing (EMNLP)}, pp.\  9275--9293, Online, November
  2020. Association for Computational Linguistics.
\newblock \doi{10.18653/v1/2020.emnlp-main.746}.
\newblock URL \url{https://aclanthology.org/2020.emnlp-main.746}.

\bibitem[Wang et~al.(2021)Wang, Chen, and Zhu]{wang2021survey}
Xin Wang, Yudong Chen, and Wenwu Zhu.
\newblock A survey on curriculum learning.
\newblock \emph{IEEE Transactions on Pattern Analysis and Machine
  Intelligence}, 44\penalty0 (9):\penalty0 4555--4576, 2021.

\bibitem[Yang et~al.(2023)Yang, Jain, and Wallace]{yang2023remove}
Jinghan Yang, Sarthak Jain, and Byron~C. Wallace.
\newblock How many and which training points would need to be removed to flip
  this prediction?
\newblock 2023.
\newblock \doi{10.48550/ARXIV.2302.02169}.
\newblock URL \url{https://arxiv.org/abs/2302.02169}.

\bibitem[Yeh et~al.(2018)Yeh, Kim, Yen, and Ravikumar]{yeh2018representer}
Chih-Kuan Yeh, Joon~Sik Kim, Ian En-Hsu Yen, and Pradeep Ravikumar.
\newblock Representer point selection for explaining deep neural networks.
\newblock In \emph{Proc. NeurIPS}, 2018.

\end{thebibliography}
\bibliographystyle{tmlr}

\appendix
\section{Appendix}

\subsection{Closed form solution for learning the parameters of \simflinear{}}\label{sec:objective_soln}

Here, we provide a closed form solution to the \simflinear{} learning objective in Equation~\ref{eq:objective}, reproduced here:
\begin{equation*}
\mathcal{O}(A, B) = \sum_{t \in \obsteps} \left(L_t(z) - \hatloss_t(z)\right)^2 + \lambda (\|A\|^2 + \|B\|^2)
\end{equation*}
First, let us concatenate parameters $A \in \mathbb{R}^n$ and $B \in \mathbb{R}^n$ into a single vector, $\textbf{w} \in \mathbb{R}^{2n}$. Our approach will then be to rewrite the objective as a standard L2-regularized multivariate linear regression problem:
\begin{equation}\label{eq:matrix_objective}
\mathcal{O}(\textbf{w}) = \|\textbf{y} - \textbf{Xw}\|^2 + \lambda \|\textbf{w}\|^2
\end{equation}
where $\textbf{y}$ is a vector, $\textbf{X}$ is a matrix, and $\lambda$ is the same L2 regularization hyperparameter as in Equation~\ref{eq:objective}. Once we have it in this form, the optimal value for $\textbf{w}$ is just the standard ridge estimator \citep{hoerl1970ridge}:
\begin{equation*}
\hat{\textbf{w}} = (\textbf{X}^\top \textbf{X} + \lambda \textbf{I})^{-1} \textbf{X}^\top \textbf{y}
\end{equation*}
where $\textbf{I}$ is an $n \times n$ identity matrix.

We now define $\textbf{X}$ and $\textbf{y}$. First, recall that $\obsteps$ denotes the subset of training steps in $\rpastset$ where we recorded the loss of test example $z$ (both before and after the training step). Let $s=1,\dots,S$ index over those steps, so that $\obsteps = \{t_s\ \text{for}\ s=1,\dots,S\}$. Now, define $\textbf{y} \in \mathbb{R}^S$ to be a vector where each entry $y_s = L_{t_s}(z)$, and $\textbf{X} = [\textbf{X}^\alpha \ \textbf{X}^\beta] \in \mathbb{R}^{S\times 2n}$ to be a matrix with submatrices $\textbf{X}^\alpha \in \mathbb{R}^{S \times n}$ and $\textbf{X}^\beta  \in \mathbb{R}^{S \times n}$. The submatrices are defined as follows:

\begin{minipage}{0.45\textwidth}
\begin{equation*}
\textbf{X}^\alpha_{s,i} = \begin{cases}
L_{t_s - 1}(z) & \text{if $i \in r_{t_s}$} \\
0 & \text{otherwise.}
\end{cases}
\end{equation*}
\end{minipage}
\begin{minipage}{0.45\textwidth}
\begin{equation*}
\textbf{X}^\beta_{s,i} = \begin{cases}
1 & \text{if $i \in r_{t_s}$} \\
0 & \text{otherwise.}
\end{cases}    
\end{equation*}
\end{minipage}

Recall that $i \in r_{t_s}$ indicates whether training example $z_i$ was consumed at training step $t_s$. With these values, one can verify that Equation~\ref{eq:matrix_objective} is equal to Equation~\ref{eq:objective}.

\subsection{Data requirements for \simflinear{}}\label{sec:sample_efficiency}

In Section~\ref{sec:data_requirements}, we studied the data requirements for \simflinear{} when the training batch size is 1. We now turn to the general case of batch size > 1.

Again, we approach this question by asking what conditions are necessary to guarantee a unique solution to the \simflinear{} training objective. We will use our observations from the previous section, where we formulated \simflinear{} as a multivariate linear regression problem (Equation~\ref{eq:matrix_objective}).

When L2 regularization is enabled ($\lambda > 0)$, the conditions for a unique solution are simple but not enlightening: Equation~\ref{eq:matrix_objective} always has a unique solution, even when zero training steps are observed.

To develop better intuitions, we will study the case where L2 regularization is disabled ($\lambda = 0$). Then, Equation~\ref{eq:matrix_objective} \textbf{has a unique solution if and only if the matrix $\textbf{X}$ has linearly independent column vectors.} Note that $\textbf{X}$ has $2n$ column vectors: $n$ column vectors comprising $\textbf{X}^\alpha$ and $n$ comprising $\textbf{X}^\beta$. Each vector has dimension $S$, the number of training steps that we observed. We will use $\textbf{X}^\alpha(i)$ to denote the $i^{th}$ column of $\textbf{X}^\alpha$ and similarly for $\textbf{X}^\beta(i)$. Finally, note that $\textbf{X}^\alpha(i)$ and $\textbf{X}^\beta(i)$ are both sparse vectors with the same sparsity pattern: the $s^{th}$ entry of each vector is zero unless example $i$ was present at training step $t_s$.

We can now describe a few conditions that determine whether $\textbf{X}$ has linearly independent columns (hence guaranteeing a unique solution):
\begin{enumerate}
    \item \textbf{We always need to observe at least $S = 2n$ training steps}. If $S < 2n$, then the column vectors could not be linearly independent, since we have more vectors than dimensions.
    \item \textbf{For every training example of interest, we need to observe at least \emph{two} training steps involving that example} (just as in the batch size 1 setting). If not, then $\textbf{X}^\alpha(i)$ would be a multiple of $\textbf{X}^\beta(i)$, creating linear dependence.
    \item \textbf{We cannot obtain a unique solution if $L_t(z)$ is constant over time} (for example if training does not change the loss). This is because column $\textbf{X}^\alpha(i)$ would be equal to column $\textbf{X}^\beta(i)$ times a scaling factor, $L_t(z)$ --- again, violating linear independence. The intuition: if the loss does not change, we cannot distinguish between additive versus multiplicative effects.
\end{enumerate}

Next, we present a manually designed training curriculum with exactly
$S=2n$ training steps, and show that this curriculum is sufficient
to obtain a unique solution. We will define a curriculum of just $n$
steps, and then repeat that exact same curriculum to obtain $2n$
steps.

To define our $n$-step curriculum, we introduce a ``batching matrix'',
$\mathbf{Q}$: it is an $n\times n$ binary matrix, where $\mathbf{Q}_{ij}=1$
if the batch at training step $i$ contains example $j$, and 0 otherwise.
Furthermore, let $k$ denote the batch size, and let us assume that
$n$ is evenly divisible by $k+1$. Then, we define $\mathbf{Q}$
to have the following block-diagonal structure:
\[
\mathbf{Q}=\begin{bmatrix}\mathbf{U} & \mathbf{0} & \cdots & \mathbf{0}\\
\mathbf{0} & \mathbf{U} & \cdots & \mathbf{0}\\
\vdots & \vdots & \ddots & \vdots\\
\mathbf{0} & \mathbf{0} & \mathbf{\cdots} & \mathbf{U}
\end{bmatrix}
\]
where the submatrix $\mathbf{U}$ is a $(k+1)\times(k+1)$ binary
matrix with the following structure:
\[
\mathbf{U}=\mathbf{1}\mathbf{1}^{\top}-\mathbf{I}=\begin{bmatrix}0 & 1 & \cdots & 1\\
1 & 0 & \cdots & 1\\
\vdots & \vdots & \ddots & \vdots\\
1 & 1 & \cdots & 0
\end{bmatrix}.
\]
where $\mathbf{1}$ denotes a vector of all ones, and $\mathbf{I}$
denotes an identity matrix. In other words, $\mathbf{U}$ is a matrix
where all non-diagonal entries are 1, and all diagonal entries are
0.

By this construction, each row of $\mathbf{Q}$ sums to $k$, our
desired batch size. Furthermore, we can show that $\mathbf{Q}$ is
full rank. First, note that $\mathbf{U}$ is full rank, because we
can use rank-preserving elementary column operations to convert it
into an identity matrix: first, we can sum all columns and divide
by $k$ to get a vector of all ones, \textbf{$\mathbf{1}$}. Then,
if we multiply each column in $\mathbf{U}$ by $-1$ and add the vector
of ones, the result is an identity matrix. Finally, recall that any
block-diagonal matrix is full rank if each sub-matrix on its diagonal
is full rank (one can apply elementary column operations to convert
each sub-matrix into an identity matrix, without affecting any other
rows or columns). Hence, $\mathbf{Q}$ is full rank.

Now that we have defined our $n$-step curriculum using $\mathbf{Q}$,
we can write $\mathbf{X}$ as follows:
\[
\text{\textbf{X}}=\begin{bmatrix}\mathbf{X}^{\alpha} & \mathbf{X}^{\beta}\end{bmatrix}=\begin{bmatrix}\text{diag}(L_{1:n})\mathbf{Q} & \mathbf{Q}\\
\text{diag}(L_{n+1:2n})\mathbf{Q} & \mathbf{Q}
\end{bmatrix}
\]
where $\text{diag}(L_{1:n})$ denotes a diagonal matrix whose diagonal entries 
are equal to $L_1(z), L_2(z), \dots, L_n(z)$.
Next, we will perform several rank-preserving transformations on $\mathbf{X}$.
First, we left-multiply by $\begin{bmatrix}\mathbf{Q}^{-1} & \mathbf{0}\\
\mathbf{0} & \mathbf{Q}^{-1}
\end{bmatrix}$:

\begin{align*}
\begin{bmatrix}\text{diag}(L_{1:n})\mathbf{Q} & \mathbf{Q}\\
\text{diag}(L_{n+1:2n})\mathbf{Q} & \mathbf{Q}
\end{bmatrix} & \rightarrow\begin{bmatrix}\text{diag}(L_{1:n}) & \mathbf{I}\\
\text{diag}(L_{n+1:2n}) & \mathbf{I}
\end{bmatrix}
\end{align*}
This is rank-preserving because the left-multiplier is full rank (it
is block-diagonal and $\mathbf{Q}$ is full rank). Then, we subtract
the lower rows from the upper rows (rank-preserving elementary row
operations):
\[
\begin{bmatrix}\text{diag}(L_{1:n}) & \mathbf{I}\\
\text{diag}(L_{n+1:2n}) & \mathbf{I}
\end{bmatrix}\rightarrow\begin{bmatrix}\text{diag}(L_{1:n}-L_{n+1:2n}) & \mathbf{0}\\
\text{diag}(L_{n+1:2n}) & \mathbf{I}
\end{bmatrix}
\]
Now, let us consider the upper-left corner, $\text{diag}(L_{1:n}-L_{n+1:2n})$.
This is a diagonal matrix, so it is full rank if $L_{t}(z) \neq L_{t+n}(z)$
for any $t$. This typically holds, since an example's loss almost always
changes after $n$ steps. Under this additional assumption, we can apply further
elementary row operations to convert the upper left corner into an
identity matrix:
\begin{align*}
\begin{bmatrix}\text{diag}(L_{1:n}-L_{n+1:2n}) & \mathbf{0}\\
\text{diag}(L_{n+1:2n}) & \mathbf{I}
\end{bmatrix} & \rightarrow\begin{bmatrix}\mathbf{I} & \mathbf{0}\\
\text{diag}(L_{n+1:2n}) & \mathbf{I}
\end{bmatrix}
\end{align*}
With identity in the upper left corner, we can apply more elementary
row operations to cancel out the lower-left corner, to obtain a full
rank identity matrix:
\[
\begin{bmatrix}\mathbf{I} & \mathbf{0}\\
\text{diag}(L_{n+1:2n}) & \mathbf{I}
\end{bmatrix}\rightarrow\begin{bmatrix}\mathbf{I} & \mathbf{0}\\
\mathbf{0} & \mathbf{I}
\end{bmatrix}
\]
Since all our operations were rank-preserving, this shows that \textbf{$\mathbf{X}$}
is full rank.

\subsection{Comparing the computational cost of \simfadd{} versus \tracincp{}}\label{sec:compute}

As noted in Section~\ref{sec:connections}, both \simfadd{} and \tracincp{} represent the same additive model of influence. The main difference is that \simfadd{} fits a model to \emph{actual observed losses}, while \tracincp{} uses \emph{hypothetical losses}.

Let $V_L$ denote the computational cost of computing the loss, $L_t(z)$, and let $V_G$ denote the cost of computing the loss gradient, $\nabla_\theta L_t(z)$. For many modern neural network implementations, $V_G \approx 2V_L$.

In \simfadd{}, we compute an actual loss reduction, $L_t(z) - L_{t+1}(z)$, which costs $2V_L$. In \tracincp{}, we compute a hypothetical loss reduction, $\nabla_\theta L_t(z)^\top \nabla_\theta L_t(z_i)$, which costs $2V_G$.

As noted in Section~\ref{sec:approach}, \simfadd{} requires us to roughly observe one loss reduction per model parameter. Each simulator has $n$ parameters, one for each training example. Furthermore, we fit a separate simulator for each test example of interest ($m$ total). Hence, to get a simulator for every test example, modeling every training example, the total cost of \simfadd{} is $2nmV_L$.

To achieve the same goal with \tracincp{}, we must compute a hypothetical loss reduction caused by every training example ($n$) on every test example ($m$), at every checkpoint ($C$). Hence, naively, the total cost of \tracincp{} is $2nmKV_G$. However, we can do better than this. Instead of computing the gradient $\nabla_\theta L_t(z)$ from scratch each time we calculate a hypothetical loss reduction, we can precompute and cache the gradient for every train and test example, just once for each checkpoint. This precomputation costs $(n + m)KV_G$. Then, let us assume that the actual dot product between gradients has negligible cost compared to computing the gradient itself. So, \tracincp{} costs approximately $(n + m)KV_G$.

Now we can compare \simfadd{}, $2nmV_L$, and \tracincp{}, $(n + m)KV_G$. Making the common assumption that $V_G = 2V_L$, we can see that the cost of the two approaches is the same when $C = nm/(n+m)$:
\begin{equation*}
    \text{\tracincp{}} = (n + m)KV_G
    = nm V_G
    = 2nm V_L
    = \text{\simfadd{}}
\end{equation*}
In the common situation where $n \gg m$, we have that $nm/(n+m) \approx m$. So the general conclusion is that \simfadd{} becomes more expensive than \tracincp{} when the number of test examples that you wish to simulate ($m$) grows larger than the number of checkpoints used by \tracincp{}.

Finally, \simfmult{} has the same cost as \simfadd{}, and \simflinear{} is just twice the cost of \simfadd{}.

\end{document}